\documentclass[11pt]{article}

\usepackage{scicite}
\usepackage{graphicx}
\usepackage{linguex}
\usepackage{xurl}
\usepackage{amsmath}
\usepackage{appendix}
\usepackage{booktabs}
\usepackage{soul}
\usepackage[all]{nowidow}

\usepackage{times}

\newenvironment{sciabstract}{%
\begin{quote} \bf}
{\end{quote}}

\title{Modeling rapid language learning by distilling Bayesian priors into artificial neural networks}

\author
{R.\ Thomas McCoy$^{1\ast}$ and Thomas L.\ Griffiths$^{2,1}$\\
\\
\normalsize{$^{1}$Department of Computer Science, Princeton University}\\
\normalsize{$^{2}$Department of Psychology, Princeton University}\\
\\
\normalsize{$^\ast$To whom correspondence should be addressed; E-mail:  tom.mccoy@princeton.edu.}
}

\date{}

\begin{document} 


\baselineskip16pt


\maketitle


\begin{sciabstract}
Humans can learn languages from remarkably little experience. Developing computational models that explain this ability has been a major challenge in cognitive science. 
Bayesian models that build in strong inductive biases---factors that guide generalization---have been successful at explaining how humans might generalize from few examples in controlled settings but are usually too restrictive to be tractably applied to more naturalistic data. By contrast, neural networks have flexible representations that allow them to learn well from naturalistic data but require many more examples than humans receive.
We show that learning from limited naturalistic data is possible with an approach that combines the strong inductive biases of a Bayesian model with the flexible representations of a neural network. 
This approach works by distilling a Bayesian model's biases into a neural network. Like a Bayesian model, the resulting system can learn formal linguistic patterns from a small number of examples. Like a neural network, it can also learn aspects of English syntax from a corpus of natural language---and it outperforms a standard neural network at acquiring the linguistic phenomena of recursion and priming.
Bridging the divide between Bayesian models and neural networks makes it possible to handle a broader range of learning scenarios than either approach can handle on its own.
\end{sciabstract}


\section{Introduction}

Across a remarkably wide range of settings, people make rich generalizations from limited experience.
This ability is particularly apparent in the case of language, making it a classic setting for debates about learning. From a small number of examples, people can learn new word meanings \cite{carey1978acquiring,bloom2002children,xu2007word}, new syntactic structures \cite{reber1967implicit,morgan1981role,culbertson2012learning,reeder2017distributional}, and new phonological rules \cite{wilson2006learning,finley2009artificial,moreton2012structure,newport2004learning}. A central challenge in cognitive science is understanding how people can infer so much about language from so little evidence \cite{goldin2003resilience,pearl2022poverty}. 
This puzzle is so extensively discussed that it has accumulated a number of different names, including the poverty of the stimulus \cite{clark2010linguistic}, Plato's problem \cite{chomsky1986knowledge}, and the logical problem of language acquisition \cite{baker1981logical}.

One popular approach for explaining rapid learning is to use probabilistic models based on Bayesian inference \cite{griffiths2010probabilistic,tenenbaum2011grow,perfors2010recursive,perfors2011learnability,odonnell2015productivity}. 
These models make strong commitments about how hypotheses are represented and selected, resulting in strong \textit{inductive biases}---factors that determine how a learner generalizes beyond its experience \cite{mitchell1997machine}. Bayesian models are thus well-suited for capturing the ability to learn from few examples. 
For instance, a recent Bayesian model introduced by Yang \& Piantadosi \cite{yang2022one} showed that it is possible to learn many important aspects of syntax from 10 or fewer examples. However, when Bayesian models are applied to larger datasets, they face significant challenges in specifying hypotheses that are flexible enough to capture the data yet remain computationally tractable.

Another influential modeling approach is to use neural networks \cite{mcclelland2010letting,rumelhart1986general,churchland1992computational}, which make few high-level commitments, giving them the flexibility to capture the nuances of realistic data. These systems represent hypotheses with matrices of numerical connection weights, and they use data-driven learning to find the best connection weights for the task at hand.
When data are plentiful, this approach is highly successful, yielding state-of-the-art systems such as the recent language model ChatGPT \cite{openai2023gpt4}. 
Nonetheless, the flexibility of neural networks is accompanied by weak inductive biases, making them perform poorly in settings with little available data.

\begin{figure*}[t]
\centering
\includegraphics[width=\linewidth]{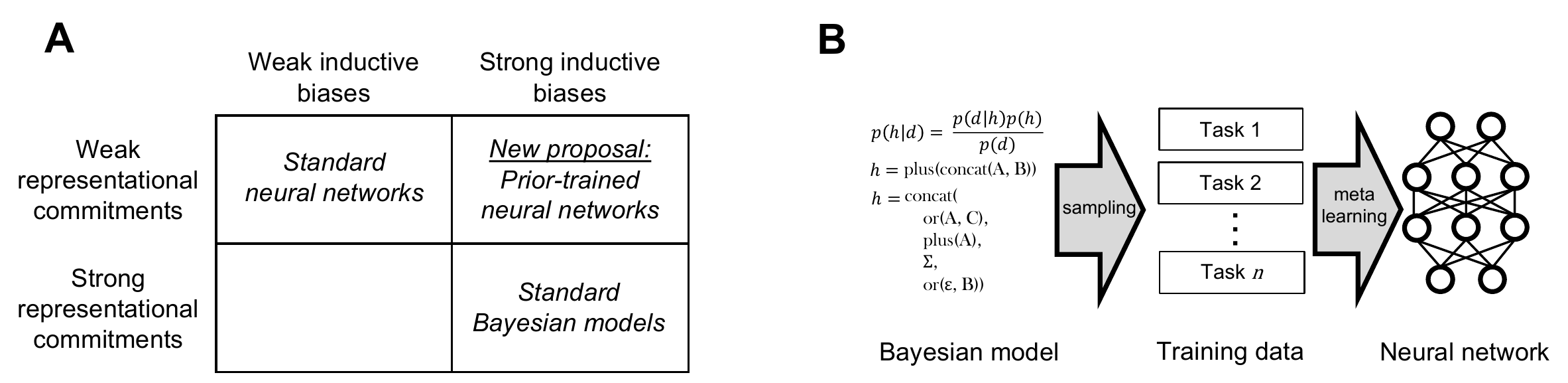}
\caption{\textbf{A.} Standard models of learning conflate strength of inductive biases with strength of representational commitments: Bayesian models have strong biases and strong representational commitments, while standard neural networks have weak biases and weak representational commitments. In this work, we create prior-trained neural networks---neural networks that have strong biases yet flexible representations. \textbf{B.} The process of inductive bias distillation that we use to give strong inductive biases to neural networks. First, a target inductive bias is instantiated in a Bayesian model which gives a prior over hypotheses. Then, hypotheses are sampled from that prior to create tasks that instantiate the inductive bias in data. Finally, a neural network meta-learns from the data, a step which transfers the inductive bias into the network.}
\label{fig:inductive_bias_distillation}
\end{figure*}

We argue that accounting for rapid learning from naturalistic data requires disentangling representations and inductive biases. 
These two factors are in principle distinct, yet historically certain types of inductive biases have always been paired with certain types of representations (Figure~\ref{fig:inductive_bias_distillation}A): 
strong inductive biases---which are important for rapid learning---have historically come with strong representational commitments (as in Bayesian models), while weak representational commitments---which provide the flexibility needed to process complex naturalistic data---have historically come with weak inductive biases (as in neural networks).
In principle, decoupling these factors would make it possible to create a system that has strong inductive biases yet weak representational commitments, enabling it---like humans---to learn rapidly without sacrificing the ability to develop more complex hypotheses. 
In practice, however, it remains far from obvious what sort of system could have both of these traits.

\renewcommand{\firstrefdash}{} 
\setlength{\Exlabelwidth}{0.7em}
\setlength{\SubExleftmargin}{1.4em}

In this work, we show how the inductive biases of a Bayesian model can be distilled into a neural network.
Our approach makes use of recent \cite{finn2017model,antoniou2018how} technical advances in \textit{meta-learning}, a machine learning technique in which a system is shown a variety of tasks from which it automatically finds an inductive bias that enables it to learn new tasks more easily \cite{schmidhuber1987evolutionary,thrun2012learning}. 
In our application of meta-learning, the tasks are sampled from a Bayesian model, thereby distilling inductive biases from the Bayesian model into the neural network. 
The result of this \textit{inductive bias distillation} is a system that has the strong inductive biases of a Bayesian model but the flexibility of a neural network. 

We use this approach to create a model of language learning.
We chose this case study because language learning is a classic problem that has long seemed to require structured symbolic representations, making it a challenging test for neural-network-based approaches.
In a setting with limited data (learning artificial formal languages from a small number of examples), our model's performance is close to that of Yang \& Piantadosi's Bayesian learner, which was the first model shown to be able to learn such languages from limited data without being substantially tailored to specific linguistic phenomena. 
Thus, even though our model is a neural network, its distilled inductive biases enable it to succeed in an environment where neural networks typically struggle, achieving a level of performance previously achieved only by a model using symbolic representations. 
In addition, the fact that our model is a neural network makes it flexible enough to handle a setting that is intractable for the Bayesian model: learning elements of English syntax from a corpus of 8.5 million words. 
Our results illustrate the possibility---and the benefits---of combining the complementary strengths of Bayesian models and neural networks.

\section{Inductive bias distillation}

As illustrated in Figure \ref{fig:inductive_bias_distillation}B, inductive bias distillation uses three steps to distill an inductive bias (called the \textit{target bias}) into a model (called the \textit{student model}). First, the target bias is defined with a Bayesian model, whose prior gives a distribution over tasks. Second, many tasks are sampled from this distribution. Finally, the student model meta-learns from these sampled tasks
to gain inductive biases that allow it to learn new tasks more easily.
By controlling the Bayesian model, we control the student model's meta-learned inductive biases.

This approach is very general: the target bias can be characterized with any distribution that can be sampled from, and the student model can be any system that is able to perform meta-learning. In our specific case, each ``task'' is a language so that the inductive bias being distilled is a prior over the space of languages \cite{wang2016galactic}. Our student model is a neural network, meaning that we distill the linguistic priors of a Bayesian model into a neural network.
This approach extends the method from our previous proof-of-concept work \cite{mccoy2020universal} by using a structured probabilistic model to define the inductive bias and by testing the model in both artificial and naturalistic scenarios.
In the rest of this section, we describe the specific form of inductive bias distillation that we perform for our language case study. 

\subsection*{Step 1: Characterizing the inductive bias}

Our starting point is the model that Yang \& Piantadosi proposed for creating a prior over formal languages \cite{yang2022one}. 
A formal language \cite{chomsky1956three,shieber1986introduction,joshi1997tree,steedman2011combinatory} is a set of strings defined by an abstract rule. For example, the set $\{A B, A B A B, A B A B A B, ...\}$ is a formal language defined by the expression $(AB)+$, meaning ``one or more copies of $A B$.'' 
The mechanisms used to define formal languages are inspired by the structure of natural language. 
The case of $(AB)+$ parallels tail recursion as seen in nested English prepositional phrases: if we take $A$ to stand for a \textbf{preposition} and $B$ to stand for a \underline{noun phrase}, then $(AB)+$ captures strings such as \textit{\textbf{under} \underline{the vase} \textbf{on} \underline{the table} \textbf{in} \underline{the library}}.
By translating linguistic structure into precise abstractions, formal languages have long facilitated mathematical analyses of language, such as proofs demonstrating which mechanisms are (in)sufficient to account for natural language syntax \cite{shieber1985evidence,frank2021variation} or phonology \cite{frank1998optimality,rogers2013cognitive}. 

In our case, the mathematical nature of formal languages makes them useful for defining a distribution over languages. 
Following the general approach used by Yang \& Piantadosi---an approach that falls within the paradigm of modeling cognition using a probabilistic language of thought \cite{goodman2008rational,piantadosi2012bootstrapping,yildirim2015learning,piantadosi2016logical,amalric2017language,rothe2017question,planton2021theory}---we specify a set of formal primitives and construct a model which probabilistically combines these primitives to create definitions of languages.
The primitives that we use are mainly drawn from standard components of regular expressions \cite{partee1993mathematical}, one particular formal language notation. Examples of these primitives include concatenation and the aforementioned recursion primitive \texttt{plus} meaning ``one or more copies of.'' An example language defined by our primitives is \texttt{concat(A, plus(C), or(F,B))}, the language of strings made of an \texttt{A} followed by one or more \texttt{C}'s followed by either \texttt{F} or \texttt{B}: $\{ACF, ACB,\allowbreak ACCF,\allowbreak ACCB,\allowbreak ACCCF, ...\}$. 
Regular expressions are limited in their expressive power: they are provably unable to capture certain aspects of natural language syntax \cite{chomsky1957syntactic}. To overcome these limitations, we augment the basic regular expression primitives in ways that increase the system's expressivity. 
See Methods (Section \ref{sec:primitives}) for a full description of our primitives.

Our complete distribution over languages is specified via a probabilistic model (structured similarly to a probabilistic context-free grammar) that defines a probability distribution over all possible combinations of our primitives. This approach assigns high probability to languages defined with few primitives and low probability to languages with more complex descriptions. 
Therefore, the  inductive bias that we aim to distill using this model is one that favors languages that can be simply expressed using our chosen primitives. 

\subsection*{Step 2: Sampling data} 

Now that we have characterized the inductive bias as a distribution over languages, the next step is to sample languages from this distribution so that the student model can meta-learn from these languages.
This step is straightforward because the distribution is defined as a generative model, which automatically allows us to sample languages from this distribution and to then sample specific strings from each language.
Although it is simple, this step is conceptually important.
It bridges the divide between our probabilistic model and our neural network by instantiating our target bias in data---something that can act as common ground between two otherwise very different models.

\subsection*{Step 3: Applying meta-learning} The final step in inductive bias distillation is to have the student model meta-learn from our sampled data in order to give it the target bias. The type of system that we use as a student model is a long short-term memory neural network (LSTM; \textit{\citen{hochreiter1997}}). LSTMs have been formally shown to be capable of processing many types of formal languages \cite{merrill2020formal}, and empirically they have been very successful in processing natural language \cite{bahdanau2015neural,kiperwasser2016simple,peters2018elmo}. We also tried using Transformers \cite{vaswani2017attention}---another type of neural network that is effective for language---but we found that the distillation was not as effective for Transformers as for LSTMs, likely because LSTMs perform better than Transformers at capturing some of the formal language mechanisms underlying our set of primitives \cite{deletang2023chomskyhierarchy}.\footnote{It is possible that future modifications to Transformers will close this gap; e.g., see \cite{yao2021self}.} 

The task that our LSTMs perform is next-word prediction \cite{elman1991distributed}, also known as language modeling: Given a sequence, the LSTM aims to predict each word in the sequence conditioned on the previous words. For example, if the sequence is $ABA$, it would be expected to first predict what the first token is ($A$); then to predict what the second token is ($B$) given that the first token is $A$; then to predict what the third token is ($A$) conditioned on the prefix $AB$; and finally to produce a special end-of-sequence token conditioned on the prefix $ABA$. 
For most languages, this task cannot be solved perfectly; e.g., in English, there are many next words that could follow $The$. Therefore, the model's prediction for the next word is a probability distribution over all possible tokens (ideally assigning the highest probability to the most likely next words).
We use the next-word prediction task because prior work has found it to be effective in teaching neural networks the grammatical properties of languages \cite{gulordava2018colorless,hu2020systematic,wilcox2022using}.

Before we describe meta-learning, it is helpful to first describe standard learning. A neural network is defined by a large number of numerical parameters, such as connection weights. In standard learning, the network starts with some initial parameter values (usually random ones). The network is then shown many examples of the target task. After each example, the network's parameters are adjusted 
such that, if the network saw the same example again, it would perform slightly better on it. After many such updates, the network should have parameter values that enable it to perform the task effectively.

Various types of meta-learning have been shown to improve the generalization abilities of neural networks \cite{lake2019compositional,nye2020learning,conklin2021meta,murty2021dreca}. 
The form of meta-learning that we use is model-agnostic meta-learning (MAML; \textit{\citen{finn2017model}}). MAML can be viewed as a way to perform hierarchical Bayesian modeling \cite{grant2018recasting}, making it a natural choice for our purpose of distilling Bayesian priors. Intuitively, in our application of MAML, a network is shown many languages and thereby learns how to learn new languages more easily. More formally, the neural network $M$ is initialized with random parameter values and then goes through many episodes of standard learning. For each episode, we sample one language $L$ from our distribution of languages and then sample two sets of sentences from this language: a training set and a test set. $M$ is then copied into a temporary model $M'$, which is trained on the training set for this language using standard (non-meta) training. The trained $M'$ is then evaluated on this language's test set. Based on the errors that $M'$ makes on this test set, the parameters of the \textit{original} model $M$ are adjusted such that, if $M$ were duplicated again into a new $M'$, the new $M'$ would learn this language more effectively. In MAML, then, what is being meta-learned is the network's parameter initialization. If MAML is successful, this initialization should encode an inductive bias that enables the model to learn any language in our distribution from relatively few examples. Because we control the distribution of languages, we also control the inductive bias that is meta-learned.
We refer to a neural network that has undergone inductive bias distillation as a \textit{prior-trained neural network} because it has been given a particular prior via training. 
Prior-training is superficially similar to another approach called pre-training, but there are important differences in what these methods achieve; see Discussion (Section \ref{sec:priorvspre}).

In inductive bias distillation, meta-learning is not a hypothesis about how humans might have come by their inductive biases. Though humans certainly perform meta-learning in at least some cases \cite{lake2017building,griffiths2019doing}, we do not claim that humans' linguistic inductive biases must arise via meta-learning, nor do we claim that these inductive biases are encoded in the form that MAML encodes them (i.e., via the initial settings of connection weights). Instead, we use meta-learning purely as a tool for creating a model that has specific inductive biases.

\section{Results}

Our goal in using inductive bias distillation is to combine the strong inductive biases of a Bayesian model with the representational flexibility of a neural network. 
To test whether our model captures the strengths of both approaches, we evaluate it in two settings: one setting that has traditionally been well-handled by Bayesian models but not neural networks, and another setting where the reverse is true. 

\subsection{Formal languages}

We first evaluate our model on its ability to learn formal languages from few examples, 
an area where Bayesian models perform well but standard neural networks perform poorly.
We use the same 56 formal languages that Yang \& Piantadosi used to evaluate their Bayesian learner. 
All of these languages were withheld from the meta-learning phase of inductive bias distillation, ensuring that our model had not encountered any of these languages before being evaluated on them.
For each evaluation language, we train our model on $n$ strings drawn from that language, with $n$ ranging from 1 to 10,000 on a logarithmic scale. 
To quantify how well the trained model has learned the intended language, we compute the model's F-score, the same metric used by Yang \& Piantadosi. The F-score quantifies how closely the set of strings assigned highest probability by the model matches the set of strings that have the highest probability in the true language (see Methods, Section \ref{sec:fscore}).

\begin{figure}[t]
\centering
\includegraphics[width=0.6\textwidth]{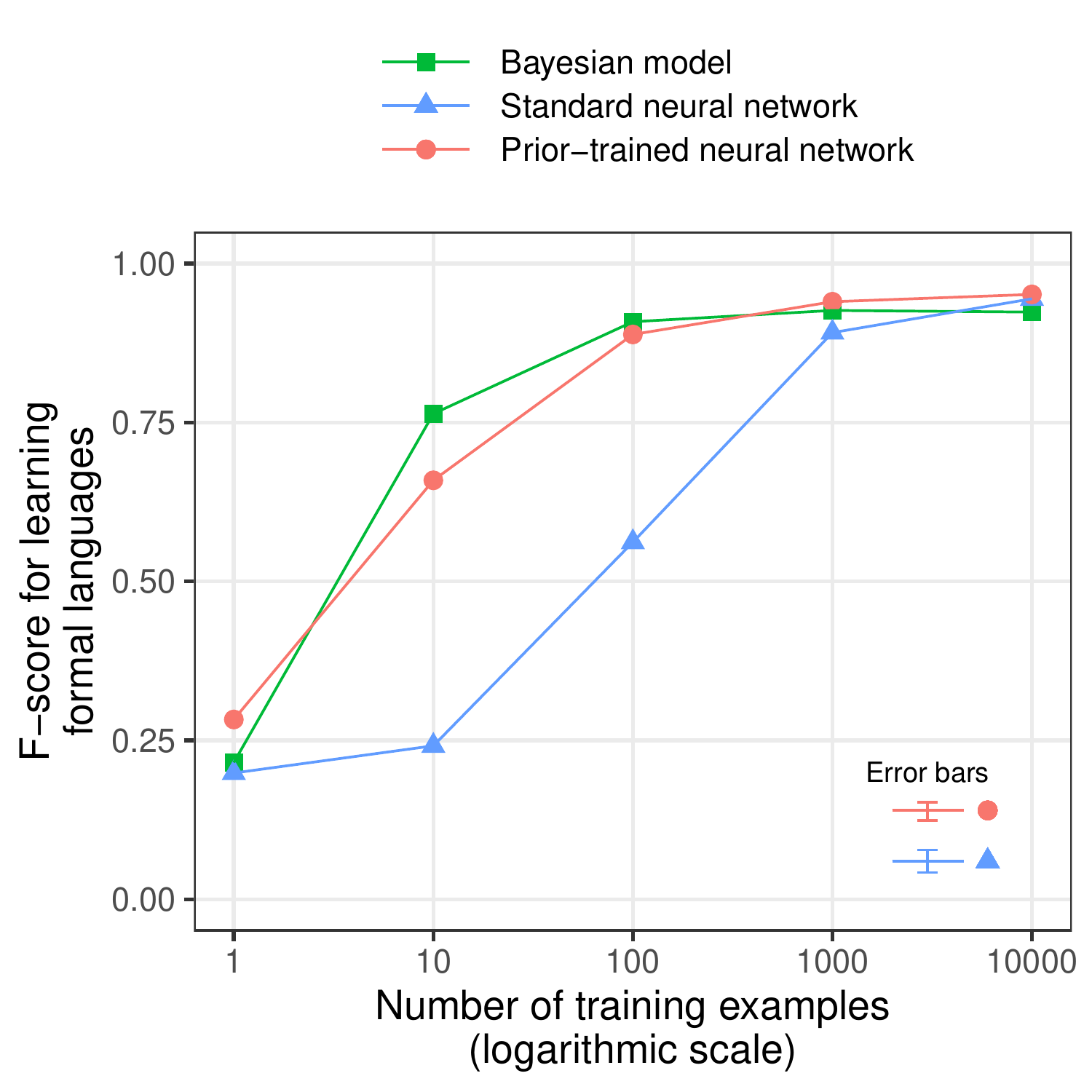}
\caption{Assessing the ability of our model to learn formal languages. This plot averages over the 56 formal languages used by Yang \& Piantadosi \cite{yang2022one}; see the supplement on the project GitHub repository for results for individual languages. The Bayesian model results are taken from Yang \& Piantadosi. The prior-trained neural network is our model that has undergone inductive bias distillation; the standard neural network has the same architecture but has not undergone distillation.
For each neural network condition, the plot shows the mean over 40 re-runs, with the bottom right showing error bars (giving the full range) averaged over the five training set sizes.} 
\label{fig:formal_language_results}
\end{figure}

This setting poses a substantial challenge for a neural network because the formal languages are defined in discrete symbolic terms. Neural networks have long been viewed as fundamentally different from symbolic processing.
Indeed, a major puzzle in cognitive science is the fact that the human mind is instantiated in a neural network yet is able to perform symbolic functions---a fact so puzzling that Smolensky \& Legendre refer to it as ``the central paradox of cognition'' \cite{smolensky2006foundational}. 
Therefore, this setting provides a challenging test of the claim that strong inductive biases can be distilled into a neural network.

Although it is a neural network, our prior-trained model displays a data efficiency similar to that of Yang \& Piantadosi's symbolic Bayesian learner (Figure \ref{fig:formal_language_results}). 
In contrast, a standard neural network is substantially more data-hungry: To reach a given level of performance, it needs about 10 times as many examples as the Bayesian learner.
The standard and prior-trained neural networks are identical in their architecture and the procedure that they use to learn a given formal language.
They only differ in that the prior-trained network has undergone inductive bias distillation while the standard one has not.
Therefore, the distillation process has succeeded at giving our model inductive biases that are useful for learning formal languages.
Although neural networks are typically associated with learning slowly, these results show that learning slowly is not a necessary aspect of being a neural network.

In addition to coming close to the Bayesian learner's data efficiency, the prior-trained network exceeds the Bayesian learner's time efficiency.
To learn one formal language, the Bayesian learner takes from 1 minute to 7 days.
Our neural network takes at most 2.5 minutes, and sometimes as little as 10 milliseconds.
The Bayesian learner is by no means slow: indeed, considering the complexity of its hypothesis space, it is blazingly fast for a learner of this sort---Yang \& Piantadosi's software package is appropriately named Fleet.
Nonetheless, the flexible parallel processing performed by neural networks enables a substantial speedup over even a very fast Bayesian learner.

\subsection{Natural language}

We next evaluate our model on its ability to learn natural language from an 8.5-million word corpus of English text \cite{yedetore2023poor}. 
This corpus, which is drawn from the CHILDES database \cite{macwhinney2000childes}, is composed of sentences that English-speaking parents spoke to their children. It is thus the type of linguistic input that humans receive when acquiring the grammatical structure of English. 
Due to the size of this dataset and the complexity of natural language, Yang \& Piantadosi's Bayesian learner cannot tractably be applied in this setting. However, the fact that our model has greater time efficiency makes processing this dataset feasible, since neural networks are well-suited for handling large naturalistic datasets, as evidenced by the success of recent large language models such as ChatGPT \cite{openai2023gpt4}.

We evaluate performance on this corpus by computing perplexity on a withheld test set. 
Perplexity is the standard metric for evaluating next-word prediction: the lower the perplexity is, the better the model is doing at predicting the next word given its context. 
A perplexity value is difficult to interpret in absolute terms, so to contextualize our model's performance we use the strong baseline of a smoothed 5-gram model (the best known non-neural system for performing next-word prediction). On this dataset, as reported in \cite{yedetore2023poor}, a smoothed 5-gram model's perplexity is 24.4.

\begin{figure*}[t]
\centering
\includegraphics[width=\linewidth]{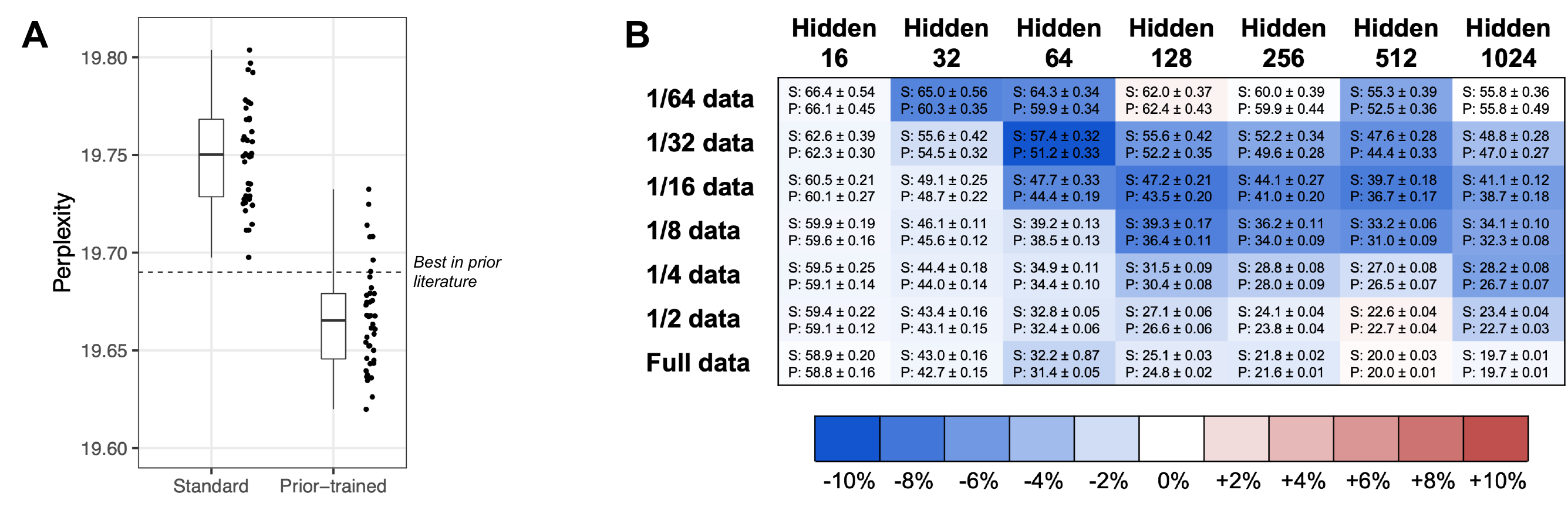}
\caption{Perplexity of neural networks trained on English text. For perplexity, lower is better. \textbf{A.} Results for our largest model size (1024 hidden units) trained on the full dataset. The boxplots show summary statistics (median, first and third quartiles, and range) across 40 runs, while the dots show the values for the individual re-runs. The dotted line is the best model from prior literature \cite{yedetore2023poor}, which was a Transformer. \textbf{B.} Effect of varying model size and amount of training data. Each cell shows the mean perplexity and a 95\% confidence interval for a standard model (S) and prior-trained model (P), across 20 runs for each model type in each cell. The shading shows the proportion by which inductive bias distillation changes the perplexity (a negative value---i.e., blue---indicates an improvement).}
\label{fig:heatmap}
\end{figure*}

Our prior-trained neural network achieves a perplexity of 19.67, substantially outperforming the 5-gram baseline. As shown in Figure \ref{fig:heatmap}A, this perplexity of 19.67 slightly improves on the perplexity of 19.75 achieved by our standard neural network ($p < 10^{-20}$), as well as the perplexity of 19.69 achieved by the best-performing neural network from prior literature \cite{yedetore2023poor}.
These results show that, despite its strong inductive biases, our model retains the flexibility needed to learn well from a naturalistic dataset.

Do our model's strong inductive biases have any human-interpretable effect on how it learns natural language? The previous paragraph might make it seem like the answer is no, since the perplexity of the prior-trained network is only slightly better than the perplexity of the standard network. 
However, even if the distilled inductive biases had a substantial effect on learning, the evaluation in the previous paragraph would be unlikely to illustrate it.
Inductive biases are what guide a learner when the training set is lacking.
In the evaluation described so far, the test set was drawn from the same distribution as the training set, and the training set was large (8.5 million words). Therefore, the training data may already give strong signals regarding how to handle the test set, leaving little room for inductive biases to affect the results. 
To better pinpoint the effects of inductive biases, we should evaluate models on situations where the training set is not as informative.
The rest of this section discusses two such situations: when the learner has access to less training data or when the learner must perform out-of-distribution generalization (generalizing to examples drawn from a different distribution than the training set).

\subsubsection{Limiting the amount of training data} 

To test whether the distilled inductive biases have a clearer effect when the amount of CHILDES training data is smaller, we trained models on varying proportions of the dataset---from one sixty-fourth of the dataset to the full dataset. 
In neural networks, data quantity can interact with model size to determine a model's performance: it is generally true that models with more parameters generalize better, but smaller models sometimes perform better when there is too little training data for a large model to learn effective values for all of its parameters. Therefore, we also varied the number of parameters by varying the hidden layer size (the size of the network's internal vector representations).

The results (Figure \ref{fig:heatmap}B) show that, in many cases, inductive bias distillation substantially improves the perplexity of models trained on English, whereas it never substantially worsens the performance. The most dramatic improvement is when models with a hidden layer size of 64 are trained on one thirty-second of the data, in which case a prior-trained network gets a perplexity of 51.2 while a standard network gets a perplexity of 57.4. The benefit remains substantial even in some larger-scale cases: with the largest hidden layer size (1024) and half of the data, a prior-trained network gets a perplexity of 22.7, compared to 23.4 for a standard network.

The full pattern of results is complex, displaying a rough diagonal band along which inductive bias distillation provides the greatest benefit: it helps most in small models with small amounts of data or in large models with large amounts of data. We conjecture that this pattern results from the interplay between two factors. 
On one hand, inductive biases are expected to be less helpful the more data there is, explaining why they have little effect in the lower left portion of the plot. On the other hand, our target inductive bias was distilled by having models meta-learn from formal languages which are superficially very different from natural language (e.g., each formal language has a vocabulary size of at most 10, while the English dataset has a vocabulary size of 17,096). 
Therefore, it presumably takes a certain amount of English data to reorient the model's distilled inductive biases from a configuration that is geared toward formal languages to one that works for natural language. The larger a model is, the more parameters it needs to adjust to perform this reorientation, potentially explaining why inductive bias distillation also has little effect in the top right corner of the plot.

\subsubsection{Testing out-of-distribution generalization}

One remarkable aspect of human language acquisition is that we learn rules for which our experience gives little or no direct evidence. Consider the following sentences. In English, a declarative sentence, such as \ref{ex:declarative_good}, can be converted to a question by replacing one of the arguments with \textit{who} and moving it to the start of the sentence, as in \ref{ex:question_good}.
There are exceptions to this general rule \cite{ross1967constraints}: it is ungrammatical to form a question of this sort when \textit{who} corresponds to a word inside a conjunction, as is the case in \ref{ex:question_bad}. 
Though situations that would give rise to \ref{ex:question_bad} are rare in standard conversation, English speakers reliably learn this constraint.

\ex. \a. The judge and the spy will visit \textbf{the banker}.\label{ex:declarative_good}
\b. \textbf{Who} will the judge and the spy visit?\label{ex:question_good}

\ex. \a. The judge will visit the spy and \textbf{the banker}.\label{ex:declarative_bad}
\b. *\textbf{Who} will the judge visit the spy and?\label{ex:question_bad}

\noindent
In addition to being an impressive aspect of acquisition, the fact that many linguistic rules are rarely illustrated in natural conversation is also an important consideration when evaluating models. The evaluation set that we have used so far is a sample of naturally-occurring text. Therefore, for many linguistic phenomena, this evaluation set likely contains few sentences for which capturing that phenomenon is important. As a consequence, a model's performance on this evaluation set does not tell us whether the model has learned the phenomena that linguists typically focus on.

To test whether models have learned particular linguistic phenomena, prior work \cite{linzen2016assessing,marvin2018targeted} has proposed an evaluation paradigm based on minimal pairs---pairs of sentences that highlight the rule being investigated. For example, if a learner recognizes that sentence \ref{ex:question_good} is better-formed than \ref{ex:question_bad}, that is evidence that the learner has learned the constraint on questions discussed above. 
The neural networks that we consider in this work are next-word prediction models, which assign a probability to every possible sequence of words. Therefore, we can apply minimal pair evaluations to our models by seeing which sentence in the pair is assigned a higher probability by the model. We use four datasets of minimal pairs, described in Methods. Each dataset targets a number of linguistic phenomena, such as the question constraint described above. 
For this analysis, we return to the setting where both the standard and prior-trained networks achieved the best perplexity (namely, training on the full dataset with a hidden layer size of 1024).

\begin{figure*}[t!]
\centering
\includegraphics[width=\linewidth]{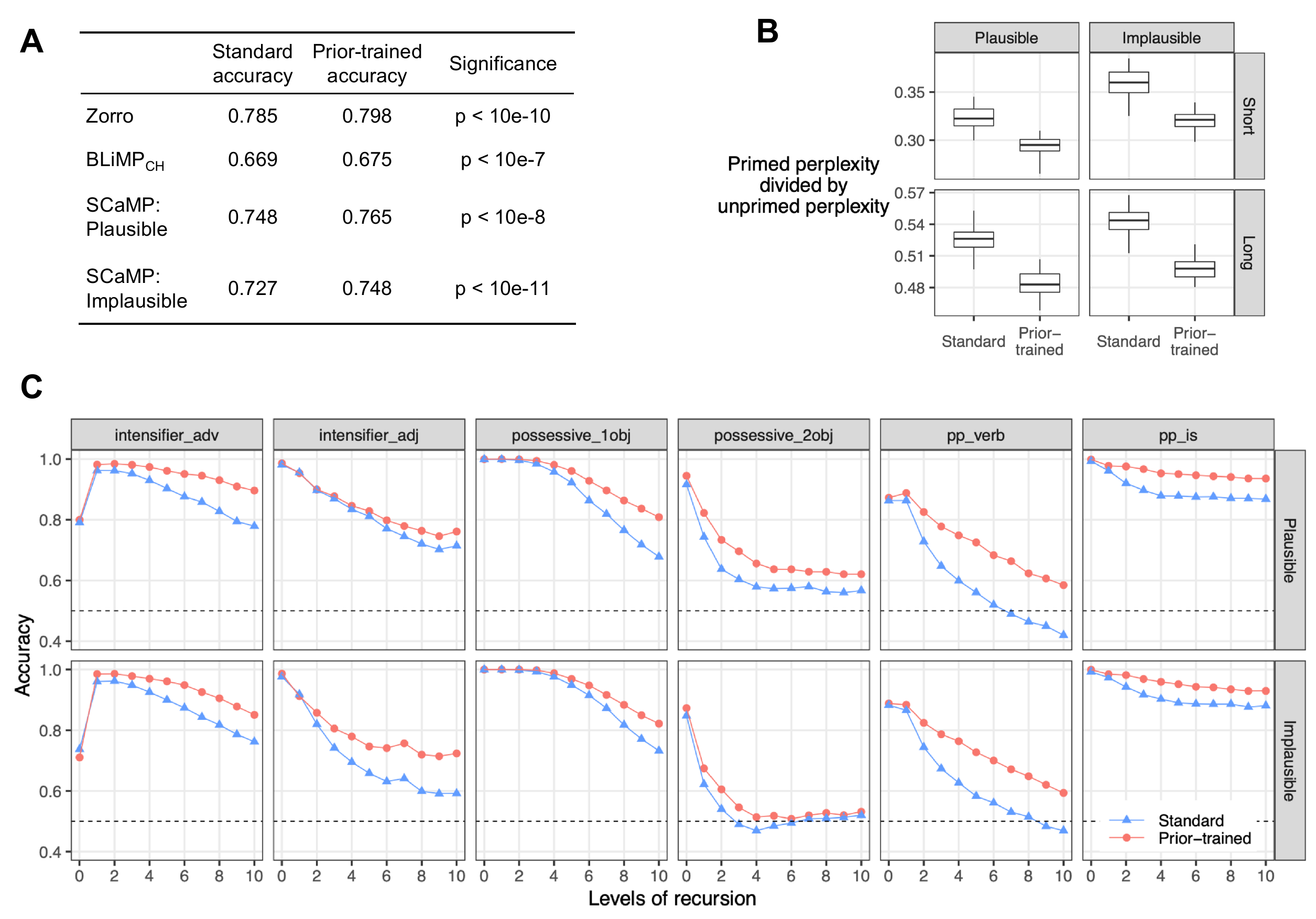}
\caption{Results on targeted linguistic evaluations. \textbf{A.} Accuracy on four minimal pair datasets that each cover a broad range of syntactic phenomena. \textbf{B.} The extent to which models display priming in sentences that are either short or long and either semantically plausible or semantically implausible. The lower the value on the y-axis is, the more extensively priming has occurred. \textbf{C.} Evaluations of recursion. The two model types score similarly when there are few levels of recursion, but at higher levels the prior-trained model often has a higher accuracy than the standard one.}
\label{fig:ood}
\end{figure*}

On all four datasets, the prior-trained neural network achieves a small but statistically significant improvement over the standard network (Figure \ref{fig:ood}A). Our online supplement\footnote{\url{https://github.com/tommccoy1/inductive-bias-distillation/blob/main/supplement/appendices_mccoy_griffiths_inductive_bias_distillation.pdf}} provides results for the individual phenomena that make up each dataset; in general, there are some phenomena where the prior-trained network substantially outperforms the standard one, but there are other phenomena where the reverse is true, and it is difficult to discern clear patterns governing which phenomena are better handled by which model (with one exception---recursion---that is discussed in the next subsection).

\subsubsection{Recursion and priming} 

The minimal pair results in the previous subsection were somewhat challenging to interpret. This fact is perhaps unsurprising because most of the phenomena tested in those evaluations did not have a clear connection to the inductive biases that we distilled. Therefore, we do not have a clear reason to expect that the distillation would help or hurt on those phenomena. In this section, we now consider two phenomena that connect more clearly to our target bias: recursion and priming.

One of the primitives we used---the \texttt{plus} primitive---enables syntactic recursion by allowing units to be repeated unboundedly many times. For instance, \texttt{plus(AB)} describes the set of strings containing one or more copies of \texttt{AB}: $\{AB, ABAB, ABABAB, ...\}$. We therefore might expect that our distilled inductive bias should improve how models process recursion in English, such as handling multiple intensifiers (\textit{the mountain is \underline{very} \underline{very} \underline{very} tall}) or possessives (\textit{my \underline{cousin's} \underline{friend's} \underline{sister's} neighbor}).\footnote{Some authors distinguish two types of repetition---recursion and iteration---based on hypothesizing that different mechanisms are used to produce the relevant sentences \cite{reich1969finiteness,christiansen1992non}. In this work, we discuss only surface strings, not the algorithms used to produce them, and therefore consider both types of repetition together under the heading of \textit{recursion}.}

Two of the minimal pair evaluation sets (SCaMP: Plausible and SCaMP: Implausible) contain stimuli that target recursion, such as the examples below. Each stimulus contains a pair of sentences that end in the same way (underlined), but where the underlined portion is a valid ending for the sentence in one example (the first example in each pair) but not the other. We compute the probability that each model assigns to the underlined portions; the model is said to be correct if it assigns higher probability to the valid case than the invalid one. Each pair involves some degree of recursion (in the examples below, each level adds an additional prepositional phrase). If a model handles recursion well, its accuracy should not degrade much when more levels of recursion are added.

\ex. \textbf{1 level:}
\a. The book on the chair \underline{is blue}.
\b. *The book was on the chair \underline{is blue}.

\ex. \textbf{2 levels:}
\a. The book on the chair in the room \underline{is blue}.
\b. *The book was on the chair in the room \underline{is blue}.

\ex. \textbf{3 levels:}
\a. The book on the chair in the room by the kitchen \underline{is blue}.
\b. *The book was on the chair in the room by the kitchen \underline{is blue}.

\noindent
Across most of our twelve recursion evaluations, the prior-trained network handles deep recursion better than the standard network (Figure \ref{fig:ood}C), supporting the hypothesis that the distilled inductive bias helps models learn recursion in English. Indeed, the recursion subsets of the SCaMP datasets are largely responsible for why the prior-trained network outperforms the standard network on these datasets in Figure~\ref{fig:ood}A. When the recursion subsets are excluded, the SCaMP$_\text{plausible}$ scores become 0.733 for the prior-trained network and 0.732 for the standard network ($p = 0.54$), and the SCaMP$_\text{implausible}$ scores become 0.721 for the prior-trained network and 0.712 for the standard network ($p < 10^{-7}$).

The other primitive that we consider here is our synchrony primitive, which enables multiple parts of a sequence to be synchronized. Most relevantly for our analysis, this primitive can capture a formal language where each sequence contains two copies of some string---e.g., \texttt{ACCDACCD} or \texttt{BDABDA}. English does not have any such patterns in the syntax of individual sentences, but this type of pattern does occur in neighboring pairs of sentences: in our corpus, 2.8\% of sentences are identical to the sentence before them. (Recall that our corpus contains sentences spoken by parents to their children; apparently, parents commonly repeat sentences). For instance, the first 6 sentences in the corpus are:

\ex. you had a what ?\\
the basketball had eyes ?\\
the basketball had eyes ?\\
eyes ?\\
you're very excited .\\
you're very excited .

\noindent
Such tendencies are not just statistical properties of corpora; they are also leveraged by language users during sentence processing, as evidenced by  priming---the tendency for language users to produce \cite{bock1986syntactic,mahowald2016meta} and expect \cite{arai2007priming,schuster2020know} sentences that are similar to others they have recently encountered. Like humans, neural network language models also display priming effects \cite{vanschijndel2018neural,prasad2019using,sinclair2022structural}.

Because our synchrony primitive facilitates the parallelism that underlies priming, we hypothesize that our distilled inductive bias should increase the extent to which models display priming. To test this hypothesis, we compute the perplexity that models assign to sentences (underlined) in an unprimed setting where the sentence appears in isolation, as in \ref{ex:unprimed}, and a primed setting where the sentence is preceded by a copy of itself, as in \ref{ex:primed}. The more a model undergoes priming, the more its perplexity should decrease from the unprimed setting to the primed setting.\footnote{Much literature on priming discusses which particular linguistic units prime each other; e.g., can a sentence be primed by another that has the same syntactic structure but no overlap in the specific words used \cite{arai2007priming}? Here we are interested in the more basic question of whether priming occurs at all, so we chose the setting where we can be most confident that priming should occur---having the two sentences in the primed condition be identical.}
This analysis was created to test our hypothesis about priming and is not part of any of the minimal pair datasets in Figure~\ref{fig:ood}A. 

\ex. \a. \underline{The lady helped the boy in the bank.}\label{ex:unprimed}
\b. The lady helped the boy in the bank. \underline{The lady helped} \underline{the boy in the bank.}\label{ex:primed}

We find that, across all four conditions we studied, the prior-trained neural network displays a greater degree of priming than the standard network (Figure \ref{fig:ood}B). This result supports our hypothesis that the inductive bias we have distilled predisposes models toward being primed.

\section{Discussion}

We have shown that prior-trained neural networks (created by distilling Bayesian priors into a neural network) can learn effectively from few examples or from complex naturalistic data.
Standard Bayesian models and standard neural networks are effective in only one of these settings but not both. 
Our results illustrate both the possibility and the importance of disentangling strength of inductive bias from strength of representational commitments: our models have strong inductive biases instantiated in continuous vector representations, a combination that enables them---like humans---to learn both rapidly and flexibly.

\subsection{Bridging levels of analysis}

Inductive bias distillation provides a way to bridge different levels at which cognition can be analyzed. 
Marr \cite{marr1982vision} proposed that cognitive science consider three levels of analysis: the computational level, which provides an abstract characterization of the problem that the mind solves and the solution that it uses; the algorithmic level, which describes the algorithm that the mind uses to execute this solution; and the implementation level, which describes how this algorithm is implemented. 
Bayesian models are usually meant as proposals at the computational level, characterizing the inductive biases that people have (i.e., which hypotheses do people select given which data?) but remaining agnostic about how these inductive biases are realized \cite{griffiths2012bayesians,goodman2016pragmatic,degen2023rational}. 
Neural networks instead align more with the algorithmic level (and, in some cases, the implementation level). 
Therefore, our experiments give an illustration of how inductive bias distillation can connect inductive biases proposed at the computational level to models proposed at the algorithmic level.

By transferring inductive biases from Bayesian models into neural networks, inductive bias distillation takes some initial steps toward connecting Bayesian models to the brain. 
The neural networks that we used are assuredly different in many ways from the brain, but they are at least in the same general class of computational systems (namely, parallel distributed processors), making them closer to the brain than the discrete systems which are more traditionally used as implementations of Bayesian models \cite{smolensky2006harmony}.
In addition to demonstrating \textit{that} a Bayesian model can be realized in a neural network, inductive bias distillation also provides trained models that can serve as objects of study for future work that illuminates \textit{how} strong inductive biases might be realized in a neural network such as the brain. 
A final way in which inductive bias distillation is helpful from the Bayesian perspective is that it can be used to test which algorithmic approaches are best suited for a given Bayesian model; for example, we found that our target inductive bias could be much more readily instantiated in one type of neural network (the LSTM architecture) than another (the Transformer architecture).

One major source of tension between Bayesian models and neural networks is differences in their representations:
many Bayesian models operate over discrete symbolic structures, whereas neural networks use continuous vector representations. 
Despite the seeming incompatibility between these types of representations, there is an extensive body of work showing how symbolic structures can be represented in vector space \cite{smolensky1990tensor,plate1991holographic,gayler1998multiplicative,kleyko2022vsas,smolensky2022neurocompositional}.
Our work extends this reconciliation from representations to inductive biases: We have shown that neural networks can be given an inductive bias that is defined over structured symbolic hypotheses.
Thus, the difference in representations is not an insurmountable obstacle for the goal of uniting high-level symbolic aspects of cognition with a vector-based implementation \cite{dehaene2015neural,pater2019generative,papadimitriou2020brain,dehaene2022symbols,prickett2022learning}.

Creating a single system that bridges levels of analysis has benefits that are not realized by simply having one model at each level. 
Compared to a standard neural network, a neural network with distilled inductive biases performs better when less data are available and also learns several syntactic phenomena more effectively when trained on naturally-occurring text.
Compared to a standard Bayesian model that uses symbolic representations, we found that instantiating inductive biases in a neural network results in much more time-efficient learning, which enabled us to empirically test what predictions our target inductive bias makes in a setting that is larger-scale than previously possible (namely, learning from millions of words of natural language).

\subsection{Learning from and extending Bayesian models of cognition}

In our case study, we showed how a particular inductive bias can be transferred to neural networks and improves performance in learning language. This inductive bias was inspired by the work of Yang \& Piantadosi \cite{yang2022one}, which provided a natural target for inductive bias distillation. In the more general case, however, how should we identify appropriate inductive biases to transfer to neural networks?

One valuable source of inductive biases is Bayesian models of cognition, which capture aspects of human learning by explicitly defining a prior distribution that captures human inductive biases \cite{griffiths2010probabilistic}. Tasks for meta-learning can be sampled from these priors, providing a simple route for extracting human inductive biases and transferring them into machines. Binz et al.\ \cite{binz2023meta} recently pointed out that meta-learning can be used to adapt neural networks to their environments, providing a way to extend rational models of cognition to more complex settings. Inductive bias distillation provides a complementary strategy for achieving this goal, in which we define an inductive bias through a prior distribution and then create an approximation to a rational model by distilling that prior into a neural network.

In other settings, it is far from settled which inductive biases people have.  For example, in the domain of language, there is a major debate about the biases that guide acquisition: some argue that acquisition only makes use of broad, general-purpose biases
(e.g., \textit{\citen{tomasello2003constructing,goldberg2006constructions,christiansen2008language}}),
while others posit detailed innate knowledge that is specific to language (e.g., \textit{\citen{chomsky1981lectures,crain1991language,cinque2010cartography}}).
In these cases, inductive bias distillation provides a tool that we can use to adjudicate between different hypothesized biases: We can distill a range of biases into different neural networks and then train them on the data that humans receive to see which inductive biases yield the most human-like learning. This in turn provides information about the prior distributions that make the most sense to incorporate into Bayesian models of human learning. 

\subsection{Prior-training vs.\ pre-training} \label{sec:priorvspre}

Prior-training has some superficial similarities to a popular existing approach called pre-training, in which a network is trained on a large quantity of general data before being further trained to perform some specific task, called a downstream task \cite{donahue2014decaf,mccann2017learned}. The crucial difference between prior-training and pre-training is that pre-training is a form of \textit{learning}---acquiring information that provides a head start on performing the downstream task---whereas prior-training is a form of \textit{learning to learn}---acquiring a prior that can then be used to learn to perform new tasks. Pre-training does influence a model's inductive biases \cite{papadimitriou2020learning,chan2022data,papadimitriou2023pretrain}, so it could in principle replace prior-training in our pipeline, but there are two reasons why we did not use it. First, theoretically speaking, prior-training gives us direct, controllable influence over inductive biases, shaping these biases to match those of a Bayesian model, whereas in pre-training the effects on inductive bias are indirect byproducts rather than direct targets of optimization. Second, empirically speaking, we found that pre-training did not work well for this purpose; see the supplementary material on the project GitHub.\footnote{\url{https://github.com/tommccoy1/inductive-bias-distillation/blob/main/supplement/appendices_mccoy_griffiths_inductive_bias_distillation.pdf}}

Some large-scale pre-trained models, such as ChatGPT, might be able to perform well on our evaluations, but there are several reasons why such systems are poorly suited to being models of language learning. First, they are pre-trained on large quantities of natural language, which makes them unrealistic as models of the initial state of a human language learner: Humans are not born with knowledge of particular natural languages, whereas ChatGPT and related systems do have such knowledge. This same factor would also raise challenges for interpreting the performance of ChatGPT if it did perform well; we would not be able to tell if any success it had were due to inductive biases as opposed to prior experience with the language being learned. 
Finally, as mentioned above, prior-training enables us to directly and controllably optimize a model's inductive biases, whereas the effect of pre-training on inductive biases is much more indirect and unpredictable.
One simple illustration of the difference between the two approaches can be seen in the information content of the data they involve. Our prior-training data were defined using a small number of abstract primitives, with the entire dataset being produced from a single 21-kilobyte Python file. By contrast, standard pre-training datasets contain far more information---typically hundreds of gigabytes of Internet text (a difference of about 6 orders of magnitude).

\subsection{Relevance for artificial intelligence}

Modern neural networks have revolutionized artificial intelligence \cite{lecun2015deep,sejnowski2018deep}, but there are some areas in which they still fall short. Two major remaining problems are data hunger (standard approaches require enormous training sets; \textit{\citen{linzen2020accelerate,warstadt2022artificial}}) and brittleness (even state-of-the-art systems often break when faced with examples that go beyond their training distribution; \textit{\citen{marcus1999rule,lake2018generalization,mccoy2019hans,lakretz2022transformers}}). Both of these issues could be mitigated by adjusting the inductive biases of neural networks, since inductive biases can guide a learner to learn from less data and generalize more robustly. Inductive bias distillation may help in implementing such a solution 
because it provides a way to control the inductive biases of neural networks.

It has historically been difficult to control a neural network's inductive biases because designing a neural network architecture usually amounts to specifying a set of \textit{processing} mechanisms, not a set of \textit{learning} mechanisms. 
Although processing and learning are closely connected, the connection is complicated, making it challenging to use processing mechanisms to control inductive biases. 
For example, incorporating tree-based processing mechanisms sometimes produces an inductive bias favoring tree-based hypotheses but sometimes not \cite{mccoy2020trees}.
Inductive bias distillation makes it possible to target inductive biases directly, enabling a greater degree of control than the indirect approach of influencing inductive biases via processing mechanisms.

The controllability afforded by inductive bias distillation may also enable researchers to circumvent the hardware lottery \cite{hooker2021hardware}---a hypothesized historical tendency for the performance of a modeling approach to be determined by its compatibility with current hardware rather than its in-principle strength as a model of information processing. 
By decoupling inductive bias from implementation, inductive bias distillation broadens the range of biases that are compatible with a given type of hardware. Therefore, it reduces the extent to which results are driven by implementational details, facilitating a greater focus on high-level considerations (e.g., which types of inductive inferences should learners make?). 
Most relevantly for our case study, the success of neural networks is in part due to their compatibility with graphics processing units (GPUs), which were originally developed for handling computer graphics but turned out to also be well-suited for efficiently training neural networks.
For Bayesian models, by contrast, there is not an analogous type of natural hardware to use, but distilling our Bayesian prior into a neural network allowed us to apply this prior in a way that took advantage of the efficient GPU processing that neural networks enjoy.

\subsection{Conclusion}

Humans can learn rapidly enough to acquire a word from just one or two examples, yet flexibly enough to accumulate the complex system of knowledge that characterizes a complete language.
For a model of cognition to capture both of these abilities, it would likely need to have strong inductive biases (to enable rapid learning) yet weak representational commitments (to provide the flexibility necessary for developing complex, large-scale hypotheses).
Cognitive models have typically had one but not both of these traits: Bayesian models provide strong inductive biases at the cost of making strong representational commitments, while neural networks provide flexible representations yet have weak inductive biases. 
Using inductive bias distillation, we have shown that it is possible to create a single system that has both desired traits by combining the representations of a neural network with the inductive biases of a Bayesian model.
Like a Bayesian model, the resulting system can learn formal linguistic patterns from a small number of examples; like a neural network, it can also learn aspects of English syntax from a corpus of natural language.
We hope that bridging the divide between these modeling approaches will enable us to account for both the rapidity and the flexibility of human learning.

\section{Methods}

\subsection{Formal language primitives}\label{sec:primitives} 
Our distribution over formal languages is mainly defined using standard regular expression primitives \cite{partee1993mathematical}:
\begin{itemize}
    \item Atomic alphabet symbols (\texttt{A}, \texttt{B}, ...)
    \item $\Sigma$: any symbol in the alphabet
    \item $\epsilon$: the empty string
    \item \texttt{concat}: concatenation
    \item \texttt{or}: randomly selecting one of two strings
    \item \texttt{plus}: Kleene plus, which produces one or more instances of an expression
\end{itemize}

\noindent
To overcome formal limitations in the expressive power of regular expressions \cite{chomsky1956three}, we make two enhancements to the basic regular expression primitives. 
First, the standard Kleene plus primitive enables tail recursion, in which multiple instances of an expression are joined sequentially (e.g., repeating \texttt{AB} to give \texttt{ABAB}). However, it does not enable nested recursion (also known as center embedding), in which multiple instances of an expression are nested inside each other (e.g., nesting \texttt{AB} inside \texttt{AB} to yield \texttt{AABB}). We generalize the Kleene plus by incorporating an index argument that specifies where recursed material is inserted: \texttt{plus(AB, 0, 0.5)} inserts new copies of \texttt{AB} at index 0 (the start of the string), yielding tail recursion: $\{AB, ABAB, ABABAB, ...\}$. The expression \texttt{plus(AB, 1, 0.5)} instead creates nested recursion by inserting new copies of AB in between the existing A and B: $\{AB,\allowbreak AABB,\allowbreak AAABBB, ...\}$. The final argument in this expression is the probability of continuing to insert new copies of \texttt{AB}: setting this value to 0.5 means that, in this language, the string \texttt{AB} has probability 0.5, the string \texttt{AABB} has probability $0.5 \times 0.5 = 0.25$, etc.

The second enhancement that we make to our set of primitives is the addition of a synchrony mechanism---inspired by synchronous grammars \cite{lewis1968syntax,aho1969syntax,shieber1990synchronous}---which allows different parts of a sequence to be synchronized. For example, the following defines a language in which each sequence has three parts:

\texttt{Synchrony pattern: 0,1,0}

\texttt{0: plus(or(A/B, B/D), 0, 0.5)}

\texttt{1: concat(C,C)}

\noindent
The synchrony pattern shows that the first and third parts are synchronized (with ID~0), while the middle part is independent (ID~1). The middle part is always the string \texttt{CC}. The first and third parts are sequences made of \texttt{A}, \texttt{B}, and \texttt{D}, where everywhere that there is an \texttt{A} in the first part, there is a \texttt{B} in the third part, and everywhere there is a \texttt{B} in the first part, there is a \texttt{D} in the third part. Example strings in this language include \texttt{ACCB} and \texttt{AABACCBBDB}.

With these primitives defined, we can sample a formal language by probabilistically combining the primitives to form a language description, with probabilities chosen in a way motivated by \cite{chi1999statistical}. See the supplement on GitHub for the specific probabilistic model we use for this purpose.

We use a different set of primitives from Yang \& Piantadosi because we found that, although their primitives were very effective for what Yang \& Piantadosi use them for (choosing between hypotheses), they are not well-suited for inductive bias distillation. Specifically, in inductive bias distillation, the distribution over languages is distilled into a learner by showing samples from this distribution to the learner. In a sample of 10,000 languages from Yang \& Piantadosi's prior distribution, we found that most languages were degenerate: 94.4\% contained only one unique string, and 98.7\% contained no strings with a length greater than 1. Therefore, distilling this distribution into a learner would require an unrealistically large number of samples in order to show sufficiently many examples of non-trivial languages, so we instead chose primitives that result in a greater proportion of non-trivial languages.

We tried running Yang \& Piantadosi's code with our primitives, but we found that it performed worse with these primitives than with Yang \& Piantadosi's primitives, potentially because our synchrony mechanism makes the hypothesis space difficult for their learner to search through. 
Therefore, in order to present each approach in the most favorable light possible, the results we present with Yang \& Piantadosi's model use their set of primitives; for each language, we used the highest-posterior hypothesis among the four candidates listed in their supplementary materials.

\subsection{Meta-training} For the meta-learning phase of inductive bias distillation, we used a meta-training set of 25,000 formal languages sampled from our meta-grammar, and a meta-validation set of 500 formal languages. 
Models went through one epoch on this dataset. 
For each formal language, the model was trained using stochastic gradient descent with a learning rate of 1.0 on $n$ batches of size 10 sampled from that language, where $n$ was drawn uniformly from [1,20]. We used multi-step loss \cite{antoniou2018how}: after each batch, a meta-loss term was computed based on a 1,000-item test set sampled from the formal language. 
After all batches for the language had been processed, the model then underwent a meta-update using AdamW \cite{kingma2015adam,loshchilov2018decoupled} with weight decay of 0.1 and a learning rate that began at 0, increased linearly to 0.005 during a warmup period covering the first 5\% of training, and then decayed following a single cycle (without restarts) of a cosine schedule \cite{loshchilov2017sgdr}. 
The model was evaluated on the meta-validation set after every 100 languages. The final trained model was the checkpoint with the lowest validation loss.
The model was a 2-layer LSTM \cite{hochreiter1997} with dropout \cite{srivastava2014dropout} of 0.1, weight sharing between input and output word representations \cite{press2017tied}, and a hidden layer size of 1024 (unless otherwise specified).
We also tried simply pretraining our model on the same dataset (i.e., combining all 25,000 languages into a single next-word prediction dataset), but we found that this approach performed substantially worse than using MAML; see the supplement in the project GitHub repository.
We implemented our models in PyTorch \cite{paszke2019pytorch}, with meta-training facilitated by the package \texttt{higher} \cite{grefenstette2019generalized} and some training functions based on code from the Transformers library \cite{wolf2020transformers}.

\subsection{Formal language evaluation} \label{sec:fscore}
Following Yang \& Piantadosi, we evaluate models on formal languages using the F-score defined as follows, where $S_{25}(D)$ is the 25 highest-probability strings in the formal language and $S_{25}(h)$ is the 25 highest-probability strings generated by the model: 
\begin{align}
    \text{precision} &= \frac{|S_{25}(h) \cap S(D)|}{S_{25}(h)} \\
    \text{recall} &= \frac{|S_{25}(D) \cap S(h)|}{S_{25}(D)} \\
    \text{F-score} &= 2\frac{\text{precision}\cdot\text{recall}}{\text{precision} + \text{recall}}
\end{align}

\noindent
To produce $S(h)$ and $S_{25}(h)$ from our model, we trained it on the relevant dataset then sampled 1 million sequences from it and reweighted their probabilities using a temperature of 0.5, as a measure for prioritizing the sequences that the model had the highest confidence in; for training sizes larger than 10, we also used nucleus sampling \cite{Holtzman2020curious} to truncate the distribution for each next token to the top 0.99 probability mass as another measure for reducing noise. These hyperparameters were tuned on a validation set of languages that were not in the 56-language evaluation set.

\subsection{Training on natural language} During our meta-training phase, the model only used a vocabulary size of 10, but our English corpus had a vocabulary size of 17,096. Therefore, to apply our model to English, we discarded its initial embedding layer and its final output layer, replacing them with randomly-initialized layers of the appropriate size. To select the hyperparameters for training models on this dataset, for each cell in the plot in Figure~\ref{fig:heatmap}B, we performed an extensive search over the hyperparameters of learning rate, dropout, and number of epochs. We performed this hyperparameter search separately for the prior-trained network and the standard network (using exactly the same search for each type of network, to ensure fairness), and trained each type of model using the hyperparameters that worked best for it. See the supplement on GitHub for the values of these hyperparameters.

To evaluate models on next-word prediction, we use the perplexity. Perplexity is defined as follows, where $W$ is the sequence of words being used to evalute the model, and $N$ is the length of $W$: 
\begin{align}
    \text{perplexity}(W) &= P(W)^{-\frac{1}{N}}
\end{align}

\subsection{Targeted linguistic evaluations}
The Zorro evaluation set was used unmodified from \cite{huebner2021babyberta}. The original BLiMP dataset \cite{warstadt2020blimp} included many words not present in our model's vocabulary, so we used the authors' code to regenerate the dataset using only the words in their vocabulary that appeared at least 10 times in the model's training set, resulting in the dataset we have labeled \textit{BLiMP$_{\text{CH}}$} (short for \textit{BLiMP$_{\text{CHILDES}}$}). 

We also wished to compare our model's performance on sentences that were plausible vs.\ implausible.
In the Zorro dataset, the sentences were deliberately designed to be semantically implausible, whereas the BLiMP sentences tend to be reasonably semantically plausible. However, these datasets differ in many other ways, so they cannot provide a controlled comparison on the dimension of plausibility. Instead, we generated two new datasets that are identical in structure but make different word choices to ensure a greater or lesser degree of plausibility. The result is a new dataset SCaMP (Selectional Category Minimal Pairs), which has a semantically-plausible version and a semantically-implausible version. Our additional evaluations targeting recursion and priming were generated from the same codebase as these two new minimal pair datasets.

\subsection{Data and code availability} All of our materials are publicly available on GitHub: \url{https://github.com/tommccoy1/inductive-bias-distillation}.

\section*{Acknowledgments}

For helpful discussion, we are grateful to Adele Goldberg, Erin Grant, Tal Linzen, Paul Smolensky, and the members of the Princeton Computational Cognitive Science Lab. We thank Steven Piantadosi for assistance in using the Fleet software package.
This material is based upon work supported by the National Science Foundation SBE
Postdoctoral Research Fellowship under Grant No.\ 2204152 and the Office of Naval Research under grant No.\ N00014-18-1-2873, and the experiments were conducted using computing resources managed by Princeton Research Computing. Any opinions, findings, and conclusions or recommendations expressed in this material
are those of the authors and do not necessarily reflect the views of the National Science
Foundation, the Office of Naval Research, or Princeton Research Computing.

\bibliography{scibib}

\begin{thebibliography}{100}

\bibitem{carey1978acquiring}
S.~Carey, E.~Bartlett, Acquiring a single new word, {\it Papers and Reports on
  Child Language Development\/} {\bf 15}, 17 (1978).

\bibitem{bloom2002children}
P.~Bloom, {\it How Children Learn the Meanings of Words\/} (MIT press, 2002).

\bibitem{xu2007word}
F.~Xu, J.~B. Tenenbaum, Word learning as {Bayesian} inference., {\it
  Psychological Review\/} {\bf 114}, 245 (2007).

\bibitem{reber1967implicit}
A.~S. Reber, Implicit learning of artificial grammars, {\it Journal of Verbal
  Learning and Verbal Behavior\/} {\bf 6}, 855 (1967).

\bibitem{morgan1981role}
J.~L. Morgan, E.~L. Newport, The role of constituent structure in the induction
  of an artificial language, {\it Journal of Verbal Learning and Verbal
  Behavior\/} {\bf 20}, 67 (1981).

\bibitem{culbertson2012learning}
J.~Culbertson, P.~Smolensky, G.~Legendre, Learning biases predict a word order
  universal, {\it Cognition\/} {\bf 122}, 306 (2012).

\bibitem{reeder2017distributional}
P.~A. Reeder, E.~L. Newport, R.~N. Aslin, Distributional learning of
  subcategories in an artificial grammar: Category generalization and
  subcategory restrictions, {\it Journal of Memory and Language\/} {\bf 97}, 17
  (2017).

\bibitem{wilson2006learning}
C.~Wilson, Learning phonology with substantive bias: An experimental and
  computational study of velar palatalization, {\it Cognitive Science\/} {\bf
  30}, 945 (2006).

\bibitem{finley2009artificial}
S.~Finley, W.~Badecker, Artificial language learning and feature-based
  generalization, {\it Journal of Memory and Language\/} {\bf 61}, 423 (2009).

\bibitem{moreton2012structure}
E.~Moreton, J.~Pater, Structure and substance in artificial-phonology learning,
  part {I}: Structure, {\it Language and Linguistics Compass\/} {\bf 6}, 686
  (2012).

\bibitem{newport2004learning}
E.~L. Newport, R.~N. Aslin, Learning at a distance {I}. {S}tatistical learning
  of non-adjacent dependencies, {\it Cognitive Psychology\/} {\bf 48}, 127
  (2004).

\bibitem{goldin2003resilience}
S.~Goldin-Meadow, {\it The resilience of language: What gesture creation in
  deaf children can tell us about how all children learn language\/}
  (Psychology Press, 2003).

\bibitem{pearl2022poverty}
L.~Pearl, Poverty of the stimulus without tears, {\it Language Learning and
  Development\/} {\bf 18}, 415 (2022).

\bibitem{clark2010linguistic}
A.~Clark, S.~Lappin, {\it Linguistic Nativism and the Poverty of the
  Stimulus\/} (John Wiley \& Sons, 2010).

\bibitem{chomsky1986knowledge}
N.~Chomsky, {\it Knowledge of Language: Its Nature, Origin, and Use\/}
  (Praeger, 1986).

\bibitem{baker1981logical}
C.~L. Baker, J.~J. McCarthy, eds., {\it The Logical Problem of Language
  Acquisition\/} (MIT press, 1981).

\bibitem{griffiths2010probabilistic}
T.~L. Griffiths, N.~Chater, C.~Kemp, A.~Perfors, J.~B. Tenenbaum, Probabilistic
  models of cognition: {E}xploring representations and inductive biases, {\it
  Trends in Cognitive Sciences\/} {\bf 14}, 357 (2010).

\bibitem{tenenbaum2011grow}
J.~B. Tenenbaum, C.~Kemp, T.~L. Griffiths, N.~D. Goodman, How to grow a mind:
  Statistics, structure, and abstraction, {\it Science\/} {\bf 331}, 1279
  (2011).

\bibitem{perfors2010recursive}
A.~Perfors, J.~Tenenbaum, E.~Gibson, T.~Regier, How recursive is language? {A}
  {Bayesian} exploration, {\it Recursion and Human Language\/} pp. 159--175
  (2010).

\bibitem{perfors2011learnability}
A.~Perfors, J.~B. Tenenbaum, T.~Regier, The learnability of abstract syntactic
  principles, {\it Cognition\/} {\bf 118}, 306 (2011).

\bibitem{odonnell2015productivity}
T.~J. O'Donnell, {\it Productivity and Reuse in Language: A Theory of
  Linguistic Computation and Storage\/} (MIT Press, 2015).

\bibitem{mitchell1997machine}
T.~M. Mitchell, {\it Machine Learning\/} (McGraw Hill, 1997).

\bibitem{yang2022one}
Y.~Yang, S.~T. Piantadosi, One model for the learning of language, {\it
  Proceedings of the National Academy of Sciences\/} {\bf 119} (2022).

\bibitem{mcclelland2010letting}
J.~L. McClelland, {\it et~al.\/}, Letting structure emerge: connectionist and
  dynamical systems approaches to cognition, {\it Trends in Cognitive
  Sciences\/} {\bf 14}, 348 (2010).

\bibitem{rumelhart1986general}
D.~E. Rumelhart, G.~E. Hinton, J.~L. McClelland, A general framework for
  parallel distributed processing, {\it Parallel Distributed Processing:
  Explorations in the Microstructure of Cognition\/} {\bf 1}, 45 (1986).

\bibitem{churchland1992computational}
P.~S. Churchland, T.~J. Sejnowski, {\it The Computational Brain\/} (MIT press,
  1992).

\bibitem{openai2023gpt4}
OpenAI, {GPT-4} technical report (2023).

\bibitem{finn2017model}
C.~Finn, P.~Abbeel, S.~Levine, Model-agnostic meta-learning for fast adaptation
  of deep networks, {\it Proceedings of the {International Conference on
  Machine Learning}\/} (2017).

\bibitem{antoniou2018how}
A.~Antoniou, H.~Edwards, A.~Storkey, How to train your {MAML}, {\it
  International Conference on Learning Representations\/} (2019).

\bibitem{schmidhuber1987evolutionary}
J.~Schmidhuber, Evolutionary principles in self-referential learning, Ph.D.
  thesis, Institut f{\"u}r Informatik, Technische Universit{\"a}t M{\"u}nchen
  (1987).

\bibitem{thrun2012learning}
S.~Thrun, L.~Pratt, {\it Learning to Learn\/} (Kluwer Academic Publishers,
  2012).

\bibitem{wang2016galactic}
D.~Wang, J.~Eisner, The {G}alactic {D}ependencies {T}reebanks: Getting more
  data by synthesizing new languages, {\it Transactions of the Association for
  Computational Linguistics\/} {\bf 4}, 491 (2016).

\bibitem{mccoy2020universal}
R.~T. McCoy, E.~Grant, P.~Smolensky, T.~L. Griffiths, T.~Linzen, Universal
  linguistic inductive biases via meta-learning, {\it Proceedings of the 42nd
  Annual Conference of the Cognitive Science Society\/} pp. 737--743 (2020).

\bibitem{chomsky1956three}
N.~Chomsky, Three models for the description of language, {\it IRE Transactions
  on Information Theory\/} {\bf 2}, 113 (1956).

\bibitem{shieber1986introduction}
S.~M. Shieber, {\it An Introduction to Unification-Based Approaches to
  Grammar\/} (Stanford: CSLI, 2003).

\bibitem{joshi1997tree}
A.~K. Joshi, Y.~Schabes, Tree-adjoining grammars, {\it Handbook of Formal
  Languages: Volume 3 Beyond Words\/} pp. 69--123 (1997).

\bibitem{steedman2011combinatory}
M.~Steedman, J.~Baldridge, Combinatory categorial grammar, {\it
  Non-Transformational Syntax: Formal and Explicit Models of Grammar\/} pp.
  181--224 (2011).

\bibitem{shieber1985evidence}
S.~M. Shieber, Evidence against the context-freeness of natural language, {\it
  Linguistics and Philosophy\/} {\bf 8}, 333 (1985).

\bibitem{frank2021variation}
R.~Frank, T.~Hunter, Variation in mild context-sensitivity: Derivational state
  and structural monotonicity, {\it Evolutionary Linguistic Theory\/} {\bf 3},
  181 (2021).

\bibitem{frank1998optimality}
R.~Frank, G.~Satta, {O}ptimality {T}heory and the generative complexity of
  constraint violability, {\it Computational Linguistics\/} {\bf 24}, 307
  (1998).

\bibitem{rogers2013cognitive}
J.~Rogers, {\it et~al.\/}, Cognitive and sub-regular complexity, {\it Formal
  Grammar\/} (Springer, 2013), pp. 90--108.

\bibitem{goodman2008rational}
N.~D. Goodman, J.~B. Tenenbaum, J.~Feldman, T.~L. Griffiths, A rational
  analysis of rule-based concept learning, {\it Cognitive Science\/} {\bf 32},
  108 (2008).

\bibitem{piantadosi2012bootstrapping}
S.~T. Piantadosi, J.~B. Tenenbaum, N.~D. Goodman, Bootstrapping in a language
  of thought: A formal model of numerical concept learning, {\it Cognition\/}
  {\bf 123}, 199 (2012).

\bibitem{yildirim2015learning}
I.~Yildirim, R.~A. Jacobs, Learning multisensory representations for
  auditory-visual transfer of sequence category knowledge: a probabilistic
  language of thought approach, {\it Psychonomic Bulletin \& Review\/} {\bf
  22}, 673 (2015).

\bibitem{piantadosi2016logical}
S.~T. Piantadosi, J.~B. Tenenbaum, N.~D. Goodman, The logical primitives of
  thought: Empirical foundations for compositional cognitive models., {\it
  Psychological Review\/} {\bf 123}, 392 (2016).

\bibitem{amalric2017language}
M.~Amalric, {\it et~al.\/}, The language of geometry: Fast comprehension of
  geometrical primitives and rules in human adults and preschoolers, {\it PLoS
  Computational Biology\/} {\bf 13}, e1005273 (2017).

\bibitem{rothe2017question}
A.~Rothe, B.~M. Lake, T.~Gureckis, Question asking as program generation, {\it
  Advances in Neural Information Processing Systems\/} {\bf 30} (2017).

\bibitem{planton2021theory}
S.~Planton, {\it et~al.\/}, A theory of memory for binary sequences: Evidence
  for a mental compression algorithm in humans, {\it PLoS Computational
  Biology\/} {\bf 17}, e1008598 (2021).

\bibitem{partee1993mathematical}
B.~H. Partee, A.~ter Meulen, R.~Wall, {\it Mathematical Methods in
  Linguistics\/} (Kluwer, 1993).

\bibitem{chomsky1957syntactic}
N.~Chomsky, {\it Syntactic Structures\/} (Mouton de Gruyter, 1957).

\bibitem{hochreiter1997}
S.~Hochreiter, J.~Schmidhuber, Long short-term memory, {\it Neural
  Computation\/} {\bf 9}, 1735 (1997).

\bibitem{merrill2020formal}
W.~Merrill, {\it et~al.\/}, A formal hierarchy of {RNN} architectures, {\it
  Proceedings of the 58th Annual Meeting of the Association for Computational
  Linguistics\/} (Association for Computational Linguistics, Online, 2020), pp.
  443--459.

\bibitem{bahdanau2015neural}
D.~Bahdanau, K.~Cho, Y.~Bengio, Neural machine translation by jointly learning
  to align and translate, {\it International Conference on Learning
  Representations\/} (2015).

\bibitem{kiperwasser2016simple}
E.~Kiperwasser, Y.~Goldberg, Simple and accurate dependency parsing using
  bidirectional {LSTM} feature representations, {\it Transactions of the
  Association for Computational Linguistics\/} {\bf 4}, 313 (2016).

\bibitem{peters2018elmo}
M.~E. Peters, {\it et~al.\/}, Deep contextualized word representations, {\it
  Proceedings of the 2018 Conference of the North {A}merican Chapter of the
  Association for Computational Linguistics: Human Language Technologies,
  Volume 1 (Long Papers)\/} (Association for Computational Linguistics, New
  Orleans, Louisiana, 2018), pp. 2227--2237.

\bibitem{vaswani2017attention}
A.~Vaswani, {\it et~al.\/}, Attention is all you need, {\it Advances in Neural
  Information Processing Systems\/} {\bf 30}, 5998 (2017).

\bibitem{deletang2023chomskyhierarchy}
G.~Deletang, {\it et~al.\/}, Neural networks and the {C}homsky hierarchy, {\it
  International Conference on Learning Representations\/} (2023).

\bibitem{yao2021self}
S.~Yao, B.~Peng, C.~Papadimitriou, K.~Narasimhan, Self-attention networks can
  process bounded hierarchical languages, {\it Proceedings of the 59th Annual
  Meeting of the Association for Computational Linguistics and the 11th
  International Joint Conference on Natural Language Processing (Volume 1: Long
  Papers)\/} (Association for Computational Linguistics, Online, 2021), pp.
  3770--3785.

\bibitem{elman1991distributed}
J.~L. Elman, Distributed representations, simple recurrent networks, and
  grammatical structure, {\it Machine Learning\/} {\bf 7}, 195 (1991).

\bibitem{gulordava2018colorless}
K.~Gulordava, P.~Bojanowski, E.~Grave, T.~Linzen, M.~Baroni, Colorless green
  recurrent networks dream hierarchically, {\it Proceedings of the 2018
  Conference of the North {A}merican Chapter of the Association for
  Computational Linguistics: Human Language Technologies, Volume 1 (Long
  Papers)\/} (Association for Computational Linguistics, New Orleans,
  Louisiana, 2018), pp. 1195--1205.

\bibitem{hu2020systematic}
J.~Hu, J.~Gauthier, P.~Qian, E.~Wilcox, R.~Levy, A systematic assessment of
  syntactic generalization in neural language models, {\it Proceedings of the
  58th Annual Meeting of the Association for Computational Linguistics\/}
  (Association for Computational Linguistics, Online, 2020), pp. 1725--1744.

\bibitem{wilcox2022using}
E.~G. Wilcox, R.~Futrell, R.~Levy, Using computational models to test syntactic
  learnability, {\it Linguistic Inquiry\/} pp. 1--88 (2022).

\bibitem{lake2019compositional}
B.~M. Lake, Compositional generalization through meta sequence-to-sequence
  learning, {\it Advances in Neural Information Processing Systems\/} {\bf 32}
  (2019).

\bibitem{nye2020learning}
M.~I. Nye, A.~Solar-Lezama, J.~B. Tenenbaum, B.~M. Lake, Learning compositional
  rules via neural program synthesis, {\it arXiv preprint arXiv:2003.05562\/}
  (2020).

\bibitem{conklin2021meta}
H.~Conklin, B.~Wang, K.~Smith, I.~Titov, Meta-learning to compositionally
  generalize, {\it {Proceedings of the 59th Annual Meeting of the Association
  for Computational Linguistics and the 11th International Joint Conference on
  Natural Language Processing}\/} (2021), vol.~1, pp. 3322--3335.

\bibitem{murty2021dreca}
S.~Murty, T.~B. Hashimoto, C.~Manning, {DR}e{C}a: A general task augmentation
  strategy for few-shot natural language inference, {\it Proceedings of the
  2021 Conference of the North American Chapter of the Association for
  Computational Linguistics: Human Language Technologies\/} (Association for
  Computational Linguistics, Online, 2021), pp. 1113--1125.

\bibitem{grant2018recasting}
E.~Grant, C.~Finn, S.~Levine, T.~Darrell, T.~Griffiths, Recasting
  gradient-based meta-learning as hierarchical {B}ayes, {\it International
  Conference on Learning Representations\/} (2018).

\bibitem{lake2017building}
B.~M. Lake, T.~D. Ullman, J.~B. Tenenbaum, S.~J. Gershman, Building machines
  that learn and think like people, {\it Behavioral and Brain Sciences\/} {\bf
  40} (2017).

\bibitem{griffiths2019doing}
T.~L. Griffiths, {\it et~al.\/}, Doing more with less: meta-reasoning and
  meta-learning in humans and machines, {\it Current Opinion in Behavioral
  Sciences\/} {\bf 29}, 24 (2019).

\bibitem{smolensky2006foundational}
P.~Smolensky, G.~Legendre, Foundational implications of the {ICS} architecture:
  Unification in cognitive science, {\it The Harmonic Mind\/} {\bf 1}, 99
  (2006).

\bibitem{yedetore2023poor}
A.~Yedetore, T.~Linzen, R.~Frank, R.~T. McCoy, How poor is the stimulus?
  {E}valuating hierarchical generalization in neural networks trained on
  child-directed speech, {\it arXiv preprint arXiv:2301.11462\/}  (2023).

\bibitem{macwhinney2000childes}
B.~MacWhinney, {\it The {CHILDES} project: Tools for analyzing talk\/}
  (Lawrence Erlbaum Associates, 2000).

\bibitem{ross1967constraints}
J.~R. Ross, Constraints on variables in syntax, Ph.D. thesis, Massachusetts
  Institute of Technology (1967).

\bibitem{linzen2016assessing}
T.~Linzen, E.~Dupoux, Y.~Goldberg, Assessing the ability of {LSTM}s to learn
  syntax-sensitive dependencies, {\it Transactions of the Association for
  Computational Linguistics\/} {\bf 4}, 521 (2016).

\bibitem{marvin2018targeted}
R.~Marvin, T.~Linzen, Targeted syntactic evaluation of language models, {\it
  Proceedings of the 2018 Conference on Empirical Methods in Natural Language
  Processing\/} (Association for Computational Linguistics, Brussels, Belgium,
  2018), pp. 1192--1202.

\bibitem{reich1969finiteness}
P.~A. Reich, The finiteness of natural language, {\it Language\/} pp. 831--843
  (1969).

\bibitem{christiansen1992non}
M.~H. Christiansen, The (non) necessity of recursion in natural language
  processing, {\it Proceedings of the 14th Annual Conference of the Cognitive
  Science Society\/} (1992), pp. 665--670.

\bibitem{bock1986syntactic}
J.~K. Bock, Syntactic persistence in language production, {\it Cognitive
  Psychology\/} {\bf 18}, 355 (1986).

\bibitem{mahowald2016meta}
K.~Mahowald, A.~James, R.~Futrell, E.~Gibson, A meta-analysis of syntactic
  priming in language production, {\it Journal of Memory and Language\/} {\bf
  91}, 5 (2016).

\bibitem{arai2007priming}
M.~Arai, R.~P. Van~Gompel, C.~Scheepers, Priming ditransitive structures in
  comprehension, {\it Cognitive Psychology\/} {\bf 54}, 218 (2007).

\bibitem{schuster2020know}
S.~Schuster, J.~Degen, I know what you're probably going to say: Listener
  adaptation to variable use of uncertainty expressions, {\it Cognition\/} {\bf
  203}, 104285 (2020).

\bibitem{vanschijndel2018neural}
M.~van Schijndel, T.~Linzen, A neural model of adaptation in reading, {\it
  Proceedings of the 2018 Conference on Empirical Methods in Natural Language
  Processing\/} (Association for Computational Linguistics, Brussels, Belgium,
  2018), pp. 4704--4710.

\bibitem{prasad2019using}
G.~Prasad, M.~van Schijndel, T.~Linzen, Using priming to uncover the
  organization of syntactic representations in neural language models, {\it
  Proceedings of the 23rd Conference on Computational Natural Language Learning
  (CoNLL)\/} (Association for Computational Linguistics, Hong Kong, China,
  2019), pp. 66--76.

\bibitem{sinclair2022structural}
A.~Sinclair, J.~Jumelet, W.~Zuidema, R.~Fern{\'a}ndez, Structural persistence
  in language models: Priming as a window into abstract language
  representations, {\it Transactions of the Association for Computational
  Linguistics\/} {\bf 10}, 1031 (2022).

\bibitem{marr1982vision}
D.~Marr, {\it {Vision: A Computational Approach}\/} (Freeman, San Francisco,
  CA, 1982).

\bibitem{griffiths2012bayesians}
T.~L. Griffiths, N.~Chater, D.~Norris, A.~Pouget, How the {Bayesians} got their
  beliefs (and what those beliefs actually are): {C}omment on {B}owers and
  {D}avis., {\it Psychological Bulletin\/} {\bf 138}, 415 (2012).

\bibitem{goodman2016pragmatic}
N.~D. Goodman, M.~C. Frank, Pragmatic language interpretation as probabilistic
  inference, {\it Trends in Cognitive Sciences\/} {\bf 20}, 818 (2016).

\bibitem{degen2023rational}
J.~Degen, {The Rational Speech Act Framework}, {\it Annual Review of
  Linguistics\/} {\bf 9}, 519 (2023).

\bibitem{smolensky2006harmony}
P.~Smolensky, G.~Legendre, Harmony optimization and the computational
  architecture of the mind/brain, {\it The Harmonic Mind\/} {\bf 1}, 59 (2006).

\bibitem{smolensky1990tensor}
P.~Smolensky, Tensor product variable binding and the representation of
  symbolic structures in connectionist systems, {\it Artificial Intelligence\/}
  {\bf 46}, 159 (1990).

\bibitem{plate1991holographic}
T.~Plate, Holographic reduced representations: Convolution algebra for
  compositional distributed representations, {\it Proceedings of the 12th
  International Joint Conference on Artificial Intelligence\/} (1991), pp.
  30--35.

\bibitem{gayler1998multiplicative}
R.~W. Gayler, Multiplicative binding, representation operators \& analogy, {\it
  Advances in Analogy Research\/}  (1998).

\bibitem{kleyko2022vsas}
D.~Kleyko, D.~A. Rachkovskij, E.~Osipov, A.~Rahimi, A survey on
  hyperdimensional computing aka vector symbolic architectures, part {I}:
  Models and data transformations, {\it ACM Computing Surveys\/} {\bf 55}
  (2022).

\bibitem{smolensky2022neurocompositional}
P.~Smolensky, R.~T. McCoy, R.~Fernandez, M.~Goldrick, J.~Gao,
  Neurocompositional computing: From the central paradox of cognition to a new
  generation of {AI} systems, {\it AI Magazine\/} {\bf 43}, 308 (2022).

\bibitem{dehaene2015neural}
S.~Dehaene, F.~Meyniel, C.~Wacongne, L.~Wang, C.~Pallier, The neural
  representation of sequences: from transition probabilities to algebraic
  patterns and linguistic trees, {\it Neuron\/} {\bf 88}, 2 (2015).

\bibitem{pater2019generative}
J.~Pater, Generative linguistics and neural networks at 60: Foundation,
  friction, and fusion, {\it Language\/} {\bf 95}, e41 (2019).

\bibitem{papadimitriou2020brain}
C.~H. Papadimitriou, S.~S. Vempala, D.~Mitropolsky, M.~Collins, W.~Maass, Brain
  computation by assemblies of neurons, {\it Proceedings of the National
  Academy of Sciences\/} {\bf 117}, 14464 (2020).

\bibitem{dehaene2022symbols}
S.~Dehaene, F.~Al~Roumi, Y.~Lakretz, S.~Planton, M.~Sabl{\'e}-Meyer, Symbols
  and mental programs: a hypothesis about human singularity, {\it Trends in
  Cognitive Sciences\/}  (2022).

\bibitem{prickett2022learning}
B.~Prickett, A.~Traylor, J.~Pater, Learning reduplication with a neural network
  that lacks explicit variables, {\it Journal of Language Modelling\/} {\bf
  10}, 1 (2022).

\bibitem{binz2023meta}
M.~Binz, {\it et~al.\/}, Meta-learned models of cognition, {\it arXiv preprint
  arXiv:2304.06729\/}  (2023).

\bibitem{tomasello2003constructing}
M.~Tomasello, {\it Constructing a Language\/} (Harvard University Press, 2003).

\bibitem{goldberg2006constructions}
A.~E. Goldberg, {\it Constructions at Work: The Nature of Generalization in
  Language\/} (Oxford University Press, 2006).

\bibitem{christiansen2008language}
M.~H. Christiansen, N.~Chater, Language as shaped by the brain, {\it
  {Behavioral and Brain Sciences}\/} {\bf 31}, 489 (2008).

\bibitem{chomsky1981lectures}
N.~Chomsky, {\it Lectures on Government and Binding\/} (Foris, 1981).

\bibitem{crain1991language}
S.~Crain, Language acquisition in the absence of experience, {\it Behavioral
  and Brain Sciences\/} {\bf 14}, 597 (1991).

\bibitem{cinque2010cartography}
G.~Cinque, L.~Rizzi, The cartography of syntactic structures, {\it Oxford
  Handbook of Linguistic Analysis\/}  (2010).

\bibitem{donahue2014decaf}
J.~Donahue, {\it et~al.\/}, {DeCAF}: A deep convolutional activation feature
  for generic visual recognition, {\it International Conference on Machine
  Learning\/} (PMLR, 2014), pp. 647--655.

\bibitem{mccann2017learned}
B.~McCann, J.~Bradbury, C.~Xiong, R.~Socher, Learned in translation:
  Contextualized word vectors, {\it Advances in Neural Information Processing
  Systems\/} {\bf 30} (2017).

\bibitem{papadimitriou2020learning}
I.~Papadimitriou, D.~Jurafsky, {L}earning {M}usic {H}elps {Y}ou {R}ead: {U}sing
  transfer to study linguistic structure in language models, {\it Proceedings
  of the 2020 Conference on Empirical Methods in Natural Language Processing
  (EMNLP)\/} (Association for Computational Linguistics, Online, 2020), pp.
  6829--6839.

\bibitem{chan2022data}
S.~Chan, {\it et~al.\/}, Data distributional properties drive emergent
  in-context learning in {T}ransformers, {\it Advances in Neural Information
  Processing Systems\/} {\bf 35}, 18878 (2022).

\bibitem{papadimitriou2023pretrain}
I.~Papadimitriou, D.~Jurafsky, Pretrain on just structure: Understanding
  linguistic inductive biases using transfer learning, {\it arXiv preprint
  arXiv:2304.13060\/}  (2023).

\bibitem{lecun2015deep}
Y.~LeCun, Y.~Bengio, G.~Hinton, Deep learning, {\it Nature\/} {\bf 521}, 436
  (2015).

\bibitem{sejnowski2018deep}
T.~J. Sejnowski, {\it The Deep Learning Revolution\/} (MIT press, 2018).

\bibitem{linzen2020accelerate}
T.~Linzen, How can we accelerate progress towards human-like linguistic
  generalization?, {\it Proceedings of the 58th Annual Meeting of the
  Association for Computational Linguistics\/} (Association for Computational
  Linguistics, Online, 2020), pp. 5210--5217.

\bibitem{warstadt2022artificial}
A.~Warstadt, S.~R. Bowman, What artificial neural networks can tell us about
  human language acquisition, {\it Algebraic Structures in Natural Language\/}
  (CRC Press, 2022), pp. 17--60.

\bibitem{marcus1999rule}
G.~F. Marcus, S.~Vijayan, S.~Bandi~Rao, P.~M. Vishton, Rule learning by
  seven-month-old infants, {\it Science\/} {\bf 283}, 77 (1999).

\bibitem{lake2018generalization}
B.~Lake, M.~Baroni, Generalization without systematicity: On the compositional
  skills of sequence-to-sequence recurrent networks, {\it International
  Conference on Machine Learning\/} (PMLR, 2018), pp. 2873--2882.

\bibitem{mccoy2019hans}
R.~T. McCoy, E.~Pavlick, T.~Linzen, Right for the wrong reasons: Diagnosing
  syntactic heuristics in natural language inference, {\it Proceedings of the
  57th Annual Meeting of the Association for Computational Linguistics\/}
  (Association for Computational Linguistics, Florence, Italy, 2019), pp.
  3428--3448.

\bibitem{lakretz2022transformers}
Y.~Lakretz, T.~Desbordes, D.~Hupkes, S.~Dehaene, Can {T}ransformers process
  recursive nested constructions, like humans?, {\it Proceedings of the 29th
  International Conference on Computational Linguistics\/} (International
  Committee on Computational Linguistics, Gyeongju, Republic of Korea, 2022),
  pp. 3226--3232.

\bibitem{mccoy2020trees}
R.~T. McCoy, R.~Frank, T.~Linzen, Does syntax need to grow on trees? {S}ources
  of hierarchical inductive bias in sequence-to-sequence networks, {\it
  Transactions of the Association for Computational Linguistics\/} {\bf 8}, 125
  (2020).

\bibitem{hooker2021hardware}
S.~Hooker, The hardware lottery, {\it Communications of the ACM\/} {\bf 64}, 58
  (2021).

\bibitem{lewis1968syntax}
P.~M. Lewis, R.~E. Stearns, Syntax-directed transduction, {\it Journal of the
  ACM (JACM)\/} {\bf 15}, 465 (1968).

\bibitem{aho1969syntax}
A.~V. Aho, J.~D. Ullman, Syntax directed translations and the pushdown
  assembler, {\it Journal of Computer and System Sciences\/} {\bf 3}, 37
  (1969).

\bibitem{shieber1990synchronous}
S.~M. Shieber, Y.~Schabes, Synchronous {T}ree-{A}djoining {G}rammars, {\it
  {COLING} 1990 Volume 3: Papers presented to the 13th International Conference
  on Computational Linguistics\/} (1990).

\bibitem{chi1999statistical}
Z.~Chi, Statistical properties of probabilistic context-free grammars, {\it
  Computational Linguistics\/} {\bf 25}, 131 (1999).

\bibitem{kingma2015adam}
D.~Kingma, J.~Ba, Adam: A method for stochastic optimization, {\it
  International Conference on Learning Representations\/} (2015).

\bibitem{loshchilov2018decoupled}
I.~Loshchilov, F.~Hutter, Decoupled weight decay regularization, {\it
  International Conference on Learning Representations\/} (2019).

\bibitem{loshchilov2017sgdr}
I.~Loshchilov, F.~Hutter, {SGDR}: Stochastic gradient descent with warm
  restarts, {\it International Conference on Learning Representations\/}
  (2017).

\bibitem{srivastava2014dropout}
N.~Srivastava, G.~Hinton, A.~Krizhevsky, I.~Sutskever, R.~Salakhutdinov,
  Dropout: A simple way to prevent neural networks from overfitting, {\it
  Journal of Machine Learning Research\/} {\bf 15}, 1929 (2014).

\bibitem{press2017tied}
O.~Press, L.~Wolf, Using the output embedding to improve language models, {\it
  Proceedings of the 15th Conference of the {E}uropean Chapter of the
  Association for Computational Linguistics: Volume 2, Short Papers\/}
  (Association for Computational Linguistics, 2017), pp. 157--163.

\bibitem{paszke2019pytorch}
A.~Paszke, {\it et~al.\/}, {PyTorch}: An imperative style, high-performance
  deep learning library, {\it Advances in Neural Information Processing
  Systems\/} {\bf 32} (2019).

\bibitem{grefenstette2019generalized}
E.~Grefenstette, {\it et~al.\/}, Generalized inner loop meta-learning, {\it
  arXiv preprint arXiv:1910.01727\/}  (2019).

\bibitem{wolf2020transformers}
T.~Wolf, {\it et~al.\/}, Transformers: State-of-the-art natural language
  processing, {\it Proceedings of the 2020 Conference on Empirical Methods in
  Natural Language Processing: System Demonstrations\/} (Association for
  Computational Linguistics, Online, 2020), pp. 38--45.

\bibitem{Holtzman2020curious}
A.~Holtzman, J.~Buys, L.~Du, M.~Forbes, Y.~Choi, The curious case of neural
  text degeneration, {\it International Conference on Learning
  Representations\/} (2020).

\bibitem{huebner2021babyberta}
P.~A. Huebner, E.~Sulem, C.~Fisher, D.~Roth, {B}aby{BERT}a: Learning more
  grammar with small-scale child-directed language, {\it Proceedings of the
  25th Conference on Computational Natural Language Learning\/} (Association
  for Computational Linguistics, Online, 2021), pp. 624--646.

\bibitem{warstadt2020blimp}
A.~Warstadt, {\it et~al.\/}, {BL}i{MP}: The benchmark of linguistic minimal
  pairs for {E}nglish, {\it Transactions of the Association for Computational
  Linguistics\/} {\bf 8}, 377 (2020).

\end{thebibliography}

\bibliographystyle{Science}

\end{document}